\RequirePackage{fix-cm}
\documentclass[twocolumn]{svjour3}          
\smartqed  
\usepackage{graphicx}
\usepackage{hyperref}
\begin{document}

\title{Frequency-domain Learning for Volumetric-based 3D Data Perception}


\author{Zifan Yu         \and
        Suya You \and
        Fengbo Ren
}


\institute{Zifan Yu \at Arizona State University \\
              \email{zifanyu@asu.edu}           
           \and
           Suya You \at U.S. Army Research Laboratory \\
              \email{suya.you.civ@mail.mil }
              \and
        Fengbo Ren \at Arizona State University \\
        \email{renfengbo@asu.edu}
}

\date{Received: date / Accepted: date}

\maketitle

\begin{abstract}
 Frequency-domain learning draws attention due to its superior tradeoff between inference accuracy and input data size. Frequency-domain learning in 2D computer vision tasks has shown that 2D convolutional neural networks (CNN) have a stationary spectral bias towards low-frequency channels so that high-frequency channels can be pruned with no or little accuracy degradation. However, frequency-domain learning has not been studied in the context of 3D CNNs with 3D volumetric data. In this paper, we study frequency-domain learning for volumetric-based 3D data perception to reveal the spectral bias and the accuracy-input-data-size tradeoff of 3D CNNs. Our study finds that 3D CNNs are sensitive to a limited number of critical frequency channels, especially the low-frequency channels. Experiment results show that frequency-domain learning can significantly reduce the size of volumetric-based 3D inputs (based on the spectral bias) while achieving comparable accuracy with conventional spatial-domain learning approaches. Specifically, frequency-domain learning is able to reduce the input data size by 98\% in 3D shape classification while limiting the average accuracy drop within 2\%, and by 98\% in 3D point cloud semantic segmentation with a 1.48\% mean-class accuracy improvement while limiting the mean-class IoU loss within 1.55\%. Moreover, by learning from higher-resolution 3D data (i.e., 2x of the original image in the spatial domain), frequency-domain learning improves the mean-class accuracy and mean-class IoU by 3.04\% and 0.63\%, respectively, while achieving an 87.5\% input data size reduction in 3D point cloud semantic segmentation.

\keywords{deep learning \and 3D convolutional neural network \and volumetric-based perception \and frequency-domain learning}
\end{abstract}

\begin{figure*}
\begin{center}
\includegraphics[width=\linewidth]{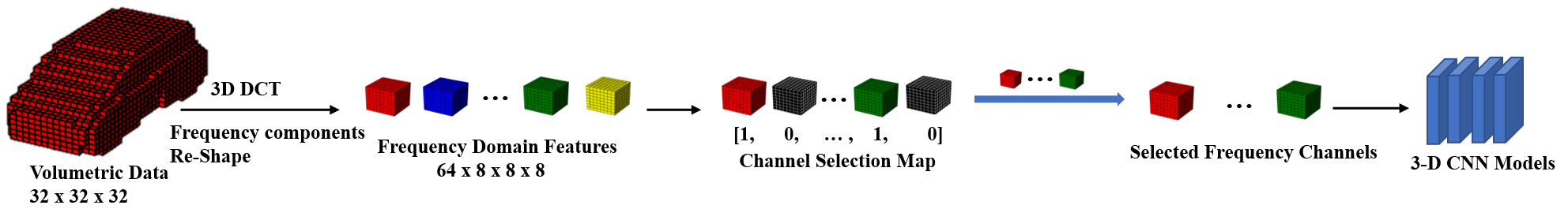}
\end{center}
   \caption{Pipeline of our 3D frequency channel selection method. The data pre-processing includes 3D DCT transformation and frequency components re-shape. The channel selection map are generated based on the spectral bias revealed by the 3D frequency domain learning. The black blocks are the muted blocks and only the selected frequency channels are the inputs for the following 3D CNN models.}
\label{fig:short}
\end{figure*}

\section{Introduction}
Frequency-domain learning draws attention from the computer vision community recently. Xu et al. \cite{2dfdl} propose a generic frequency-domain learning framework that shows a superior tradeoff between inference accuracy and input data size in various 2D computer vision tasks by learning from high-resolution inputs in the frequency domain. \cite{2dfdl} point out that the 2D CNN models have a stationary spectral bias toward a few low-frequency channels of image information in the YCbCr color space, especially the low-frequency channels of intensity (color channel Y). These observations are consistent with the theory of JPEG compression and the human visual system's bias toward low-frequency components. By removing the noisy high-frequency channels and learning from effectively higher-resolution images in the frequency domain (e.g., 2x of the original image in the spatial domain), frequency-domain learning achieves improved accuracy while the input data size in the frequency domain is equivalent to the size of low-resolution input data in the spatial domain. However, no study on frequency-domain learning has been conducted on 3D CNNs with 3D volumetric data. Thus, whether these insights can be extended to 3D CNNs and 3D data perception tasks is still in question. We hypothesize that 3D CNN models also have a certain kind of spectral bias toward a limited number of critical frequency components in various 3D data perception tasks. Thus, we conduct this study to validate our hypothesis. 

3D CNNs in 3D data perception tasks represent 3D mashes or 3D points as a dense regular 3D voxel grid \cite{3dsurvey}\cite{voxnet}. As the 3D volumetric data and 2D images are both regular grid data representations, we restrict the scope of our study to volumetric-based 3D CNN models to extend frequency-domain learning from 2D images to the 3D space. Inspired by Xu et al. \cite{2dfdl}, we first study and analyze the spectral bias of 3D CNN models toward 3D frequency-domain volumetric data by training the existing 3D CNN models (e.g., VoxNet \cite{voxnet} and VRN \cite{vrn}) with a learning-based channel selection module, which takes reshaped frequency-domain data as inputs. Our study applies limited modifications to the existing 3D CNN models (e.g., removing the downsampling layers in the first two blocks of VRN \cite{vrn} to fit the frequency-domain input data size). The spectral bias analysis results of 3D CNN models show that only a limited number of frequency channels, especially the DC channel, are highly informative to the 3D data perception tasks.  Consequently, we statically select only the critical frequency channels that have a high probability of selection to be activated for training and inference. 

A key benefit of frequency-domain learning is the ability to learn from a higher-resolution 3D representation (equivalently in the spatial domain) at a effectively reduced input data size in the frequency domain, thus providing a better tradeoff between inference accuracy and the actual input data size. Experiment results show that frequency-domain learning can significantly reduce the size of volumetric-based 3D inputs (based on the spectral bias) while achieving comparable accuracy with conventional spatial-domain learning approaches. Specifically, frequency-domain learning is able to reduce the input data size by 98\% in 3D shape classification while limiting the average accuracy drop within 2\%, and by 98\% in 3D point cloud semantic segmentation with a 1.48\% mean-class accuracy improvement while limiting the mean-class IoU loss within 1.55\%. The experiment results also show that by learning from higher-resolution 3D representation (i.e., 2x of the original 3D representation in the spatial domain), frequency-domain learning improves the mean-class accuracy and mean-class IoU by 3.04\% and 0.63\%, respectively, while achieving an 87.5\% input data size reduction in 3D point cloud semantic segmentation. 

Another benefit of frequency-domain learning with reduced input data sizes is the reduction in computational complexity and memory requirements of 3D CNN models. Our experiment results show that by learning from a 2x higher-resolution 3D representation in the frequency domain at an effectively reduced input data size, frequency-domain learning can reduce about 9\% of the floating-point operations (FLOPs) and 20\% of the GPU memory footprint required for inference compared to directly learning in the spatial domain at the same resolution. For 3D computer vision tasks like point cloud semantic segmentation, the high-resolution data has superiority in representing large-scale point clouds. But, the computation resource and memory footprint requirements also increase cubically \cite{3dsurvey}. Hence, selecting the critical and pruning the trivial frequency channels based on the learned spectral bias of 3D CNNs as a data pre-processing stage in the frequency-domain learning can potentially alleviate the large requirements for computation resources and GPU memory footprints in volumetric-based 3D vision tasks and avoid complicated modification to the existing 3D CNN models.

The contributions of this paper are as follows:

\begin{itemize}

    \item To the best of our knowledge, this is the first work that studies frequency-domain learning for 3D CNN models on 3D volumetric data. We obtain the spectral bias of 3D CNNs by training the existing 3D CNN models with a learning-based channel selection module. By analyzing the spectral bias of the 3D CNN models, we reveal that the DC components and low-frequency components already carry a significant amount of information for 3D shape classification tasks, and the top 3D frequency channels that are most informative to the point cloud semantic segmentation are more distributed across the spectrum.
    
    \item We show that the learned spectral bias of 3D CNN models can inform static channel selection to help existing 3D CNN models significantly reduce input data size with no or little accuracy degradation in shape classification and point cloud semantic segmentation. Specifically, frequency-domain learning can reduce the input data size by 98\% in shape classification, while achieving  a 0.93\% accuracy improvement on VoxNet with ModelNet and limiting the accuracy drop within 2\% on VRN with ShapeNet. In addition, frequency-domain learning can also reduce the input data size by 98\% in point cloud semantic segmentation, while achieving a 1.48\% mean-class accuracy improvement on the S3DIS dataset and limiting the mean-class IoU drop within 1.55\%.
    
    \item We investigate the impact of static channel selection in frequency-domain learning on the computational complexity and the memory footprint requirements for 3D CNN models. Our experiments show that frequency-domain learning achieves 3.04\% mean-class accuracy improvements and 0.63\% mean-class IoU improvements in learning from high-resolution data with a 9\% FLOPs decrease and a 20\% GPU memory footprint decrease compared to directly learning from high-resolution data in spatial domain.
    
\end{itemize}

\section{Related Works}

\subsection{Frequency-Domain Learning}
2D frequency-domain learning has made remarkable success in image-based computer vision tasks. 
 \cite{jpeglearning} approximately decodes the JPEG images and trains a modified ResNet \cite{resnet} with decoded DCT coefficients for the image classification task obtaining classification accuracy improvement. \cite{jpegdetection} further extends the approach in \cite{jpeglearning} to object detection task on frequency-domain features. \cite{2dfdl} reshapes the decoded DCT coefficients into a channel-wise representation and reveals the spectral bias of 2D CNN. Based on the spectral bias, they propose a frequency channel selection method with existing 2D CNN models to reduce the bandwidth required between CPUs and GPUs.  \cite{2dfdl} implements a 2D CNN-based gate module to estimate the probabilities of frequency channels to be activated by joint-training the gate module with 2D CNN models. The gate module activate channels by sampling a Bernoulli distribution based on the activation probabilities generated by CNN layers fro each channel, and \cite{2dfdl} utilizes a Gumbel Softmax trick reparameterization method to perform backpropagation through the discrete sampling.

In this paper, we extend the frequency-domain learning method proposed by \cite{2dfdl} from 2D space to 3D space to study the spectral bias of 3D CNN models. The observed spectral bias from our experiment results inspires our statical frequency channel selection method for 3D volumetric data.

\subsection{Volumetric-Based 3D Vision Methods}

Volumetric-based 3D CNN methods represent point clouds into a regular 3D grid. Then, apply 3D Convolutional Neural Network on the volumetric grid representation. \cite{voxnet} first applies a 3D CNN network on data in volumetric occupancy grid representation. The three proposed occupancy models for voxel quantization achieve close accuracy in shape classification. The binary hit model is widely utilized by following volumetric-based methods. Qi et.al \cite{subvolume} further explores the power of 3D volumetric representation and 3D CNN networks by using auxiliary training with subvolume supervision. Inspired by the high-performance 2D CNN networks (e.g, InceptionNet \cite{inception} and ResNet \cite{resnet}), \cite{vrn} proposes a VRN block with residual connection on 3D convolutional layers to make the 3D CNN networks deeper and achieves the state-of-art performance in shape classification. For point cloud segmentation using volumetric data, \cite{pointseg1} first applies a 3D convolutional neural network to label 3D points, and a point is labeled the same class with the voxel it belongs to. \cite{segcloud} improves the point cloud segmentation accuracy by adding a 3D interpolation layer to train the 3D CNN network with point-wise loss and deeper network architecture.  

In this paper, we utilize the VoxNet \cite{voxnet} and simplest VRN \cite{vrn} as our baseline methods.  \cite{vrn} presents that the ensemble of VRNs can further improve the classification accuracy. To speed the training process, we do not use an ensemble of different VRN models. We make limited modifications to VoxNet and VRN to make them fit the inputs size of data in the frequency domain but keep the network architecture unchanged. As the state-of-art volumetric-based fully CNN point cloud segmentation method proposed by \cite{segcloud} is not open-sourced, we choose to implement an encoder-decoder 3D CNN network based on the high-performance 2D CNN network \cite{rangenet} with a 3D interpolation layer \cite{segcloud} as the baseline method for point cloud segmentation.

\section{Methodology}

\subsection{Overview}
In this paper, our 3D frequency-domain learning pipeline includes 3D data pre-processing, network modifications to existing 3D CNN models, 3D CNN model spectral bias analysis, and static frequency channels selection method. As illustrated in Fig. 1, given a piece of volumetric data, it is reshaped in the 3D discrete cosine transform (DCT) domain, and the regrouped channel-wise frequency-domain volumetric features are fed into 3D CNN models for inference. The 3D CNN models are modified slightly, e.g., removing the downsampling layers in the first two VRN blocks of the VRN \cite{vrn} shape classification model, to take the 3D frequency-domain data as inputs. Then, we analyze the spectral bias of 3D CNN models from the frequency domain learning in shape classification and point cloud segmentation and verify that the reserved frequency channels contain important and sufficient information to obtain comparable classification and segmentation accuracy with models trained by full-size spatial-domain data. According to the spectral bias, we statically select specific frequency channels based on a binary channel selection map for training and inference.

\subsection{3D Data Pre-processing for Frequency-domain Learning}

 We represent the 3D data by a volumetric grid, e.g., a grid of $32 \times 32 \times 32$, and use the hit grid model proposed by \cite{voxnet} to quantify each voxel. A voxel has a binary state to indicate whether any 3D points occupy it. ``1" means occupied by at least one 3D point, and ``0" means free. Although the hit grid model loses the geometric information between closed points, many prior works \cite{voxnet}\cite{pointseg1}\cite{segcloud} show that it is enough to represent the surface and shape of objects. Except for point coordinate attributes, if a point has other attributes like mapped color information, a voxel is quantified by the average attribute value of all points that the voxel contains. 
 
 Then, the quantized volumetric data are converted to the frequency domain by $4 \times 4 \times 4$ 3D DCT transformation. A quantized volumetric data is divided into sub-blocks of size $4 \times 4 \times 4$. Each sub-block is converted to the frequency domain by applying a $4 \times 4 \times 4$ 3D DCT transformation, and one frequency-domain sub-block has 64 frequency components. The 64 components are ordered and indexed by a 3D zigzag map. All components of the same frequency in each sub-block are grouped into one channel. Therefore, for a piece of volumetric data of size $32 \times 32 \times 32$, its corresponding data in the frequency domain is of size $64 \times N_{attributes} \times 8 \times 8 \times 8$ after the above data pre-processing. $N_{attributes}$ is the number of attributes that a voxel has. For example, one voxel in the S3DIS dataset has four attributes, i.e. binary occupation stata and three-channel color information; its corresponding frequency-domain data has $256$ channels. As one channel contains frequency components from all sub-blocks, the spatial information between sub-blocks remains in each channel. Lastly, each frequency channel is normalized by the mean and variance of all training samples. 

\subsection{3D Frequency Channel Selection and 3D CNN Network Modifications}

\begin{figure}
\begin{center}
\includegraphics[width=\linewidth]{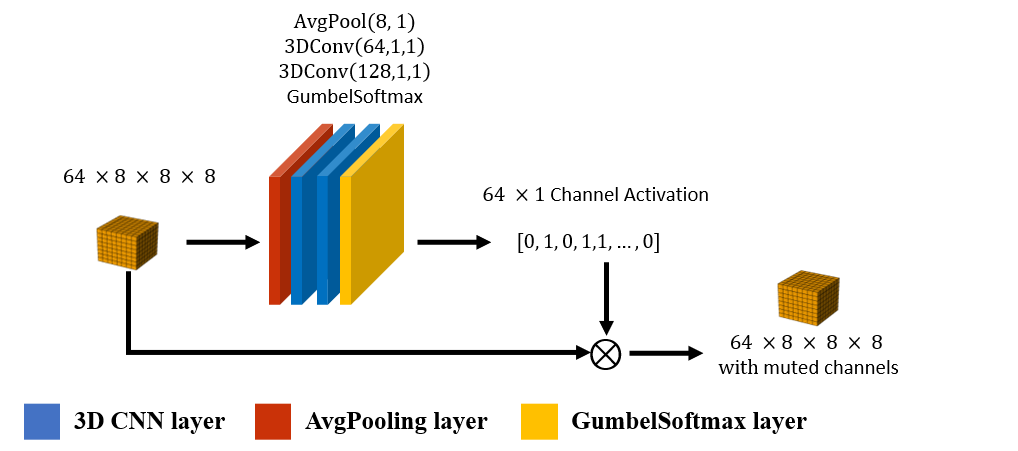}
\end{center}
   \caption{Architecture of the 3D frequency channel selection gate module. For a 3D convolutional layer $3DConv(n, k, s)$,  $n$ is the number of output channels, $k$ is the kernel size, and $s$ is the stride size. For a average pooling layer $Pool(k, s)$, k is the kernel size, and $s$ is the stride size.}
\label{fig:short}
\end{figure}

 As we hypothesis that 3D CNN models are also more sensitive to several critical frequency channels, we extend the learning-based frequency channel selection method \cite{2dfdl} to 3D CNN models to reveal the spectral bias. In this paper, we replace all 2D CNN layers and pooling layers of the gate module with 3D CNN layers and 3D pooling layers to fit the 3D frequency-domain inputs. For a frequency-domain input of size $C \times 8 \times 8 \times 8$, a global average pooling layer firstly converts it to $C \times 1 \times 1 \times 1$. Then, following two 3D CNN layers with kernel size equal to 1 generate an activated probability and a non-activated probability for each channel. After sampling a Bernoulli distribution, the output of the gate module is a binary state for each frequency channel. ``1" means this frequency channel is activated for being input for the following CNN networks, and ``0" means this channel is muted. Fig. 2 demonstrates the architecture of our 3D frequency channel selection gate module. For a channel $x_{i}$, $F(x_{i}) \in \{0, 1\}$ is its output of the gate module. The channels $x_{i}^{\prime}$ for following CNN networks is the element-wise product of $x_{i}$ and $F(x_{i})$, as shown in Equation (1).  
\begin{equation}
    x_{i}^{\prime} = x_{i} \odot F(x_{i})
\end{equation}
$\odot$ is the element-wise product. The number of selected channels is a regularization term of the loss function, and a hyperparameter $\lambda$ weights it. Therefore, for any tasks in this paper, the loss function is expressed as
\begin{equation}
    L = L_{accuracy} + \lambda \times \sum_{i}^{C} F(x_{i}),
\end{equation}

where $\L_{accuracy}$ is the loss function of 3D vision tasks, e.g. cross-entropy loss for the shape classification task. During joint-training with 3D vision tasks, we explore the spectral bias by modifying the weight of the regularization term to control the number of activated frequency channels. 

 \begin{figure*}
\begin{center}
\includegraphics[width=0.9\linewidth]{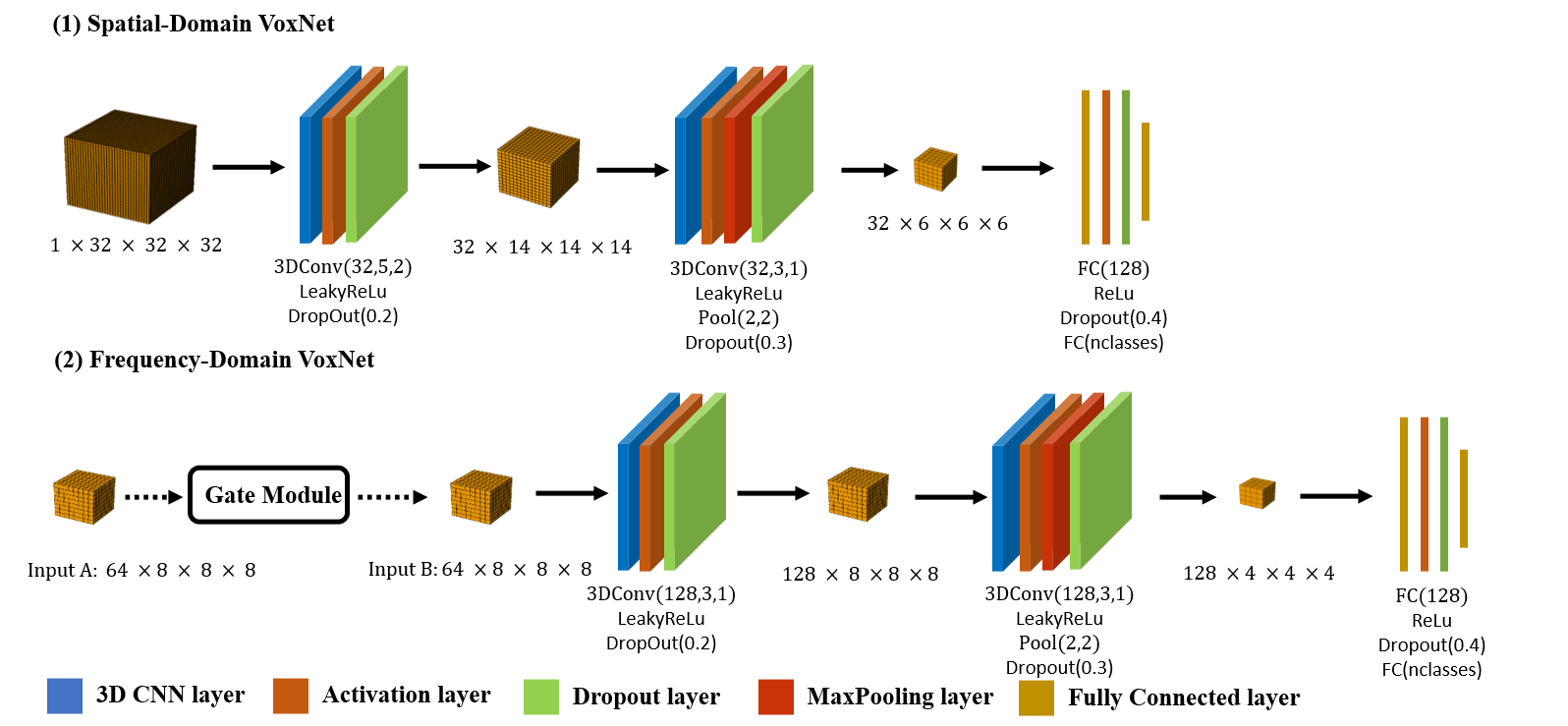}
\end{center}
   \caption{Architecture of the spatial-domain VoxNet and frequency-domain VoxNet with a gate module. The dotted lines represent the operations performed during the joint-training with the gate module to explore the 3D spectral bias. For the static channel selection method, the network training and inference start from the Input B, and the Input B has the size of $n \times 8 \times 8 \times 8$, where $n$ is the number of selected channels.}
\label{fig:short}
\end{figure*}

\begin{figure*}
\begin{center}
\includegraphics[width=\linewidth]{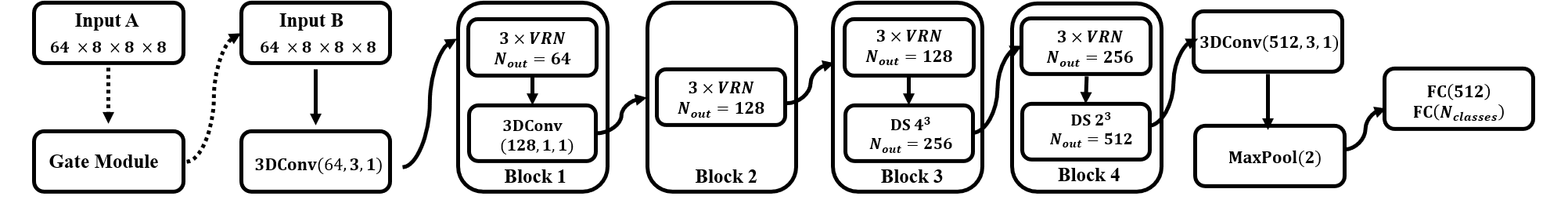}
\end{center}
   \caption{Architecture of the frequency-domain VRN. The dotted lines represent the operations performed during the joint-training with the gate module to explore the 3D spectral bias. For the static channel selection method, the network training and inference start from the Input B, and the Input B has the size of $n \times 8 \times 8 \times 8$, where $n$ is the number of selected channels.}
\label{fig:short}
\end{figure*}

 \begin{figure*}
\begin{center}
\includegraphics[width=0.9\linewidth]{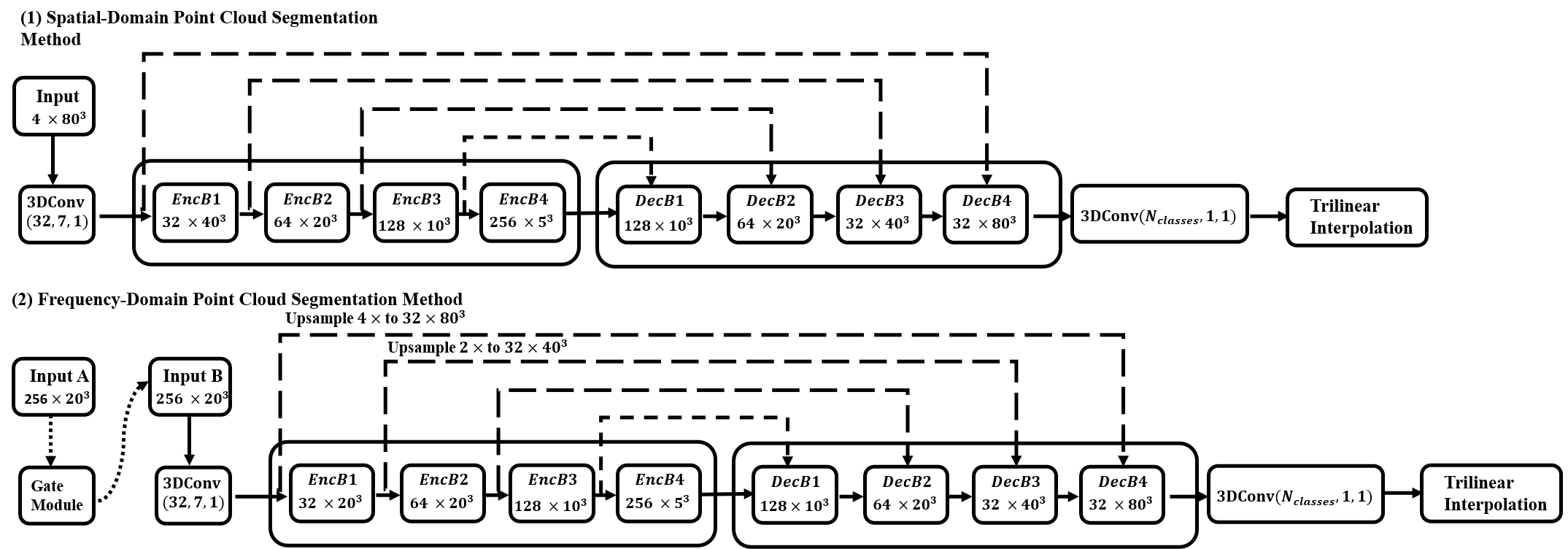}
\end{center}
   \caption{Architectures of point cloud segmentation methods.The dotted lines represent the operations performed during the joint-training with the gate module to explore the 3D spectral bias. The dash lines are skip connections from encoder blocks to decoder blocks.}
\label{fig:short}
\end{figure*}

 \begin{figure}
\begin{center}
\includegraphics[width=0.9\linewidth]{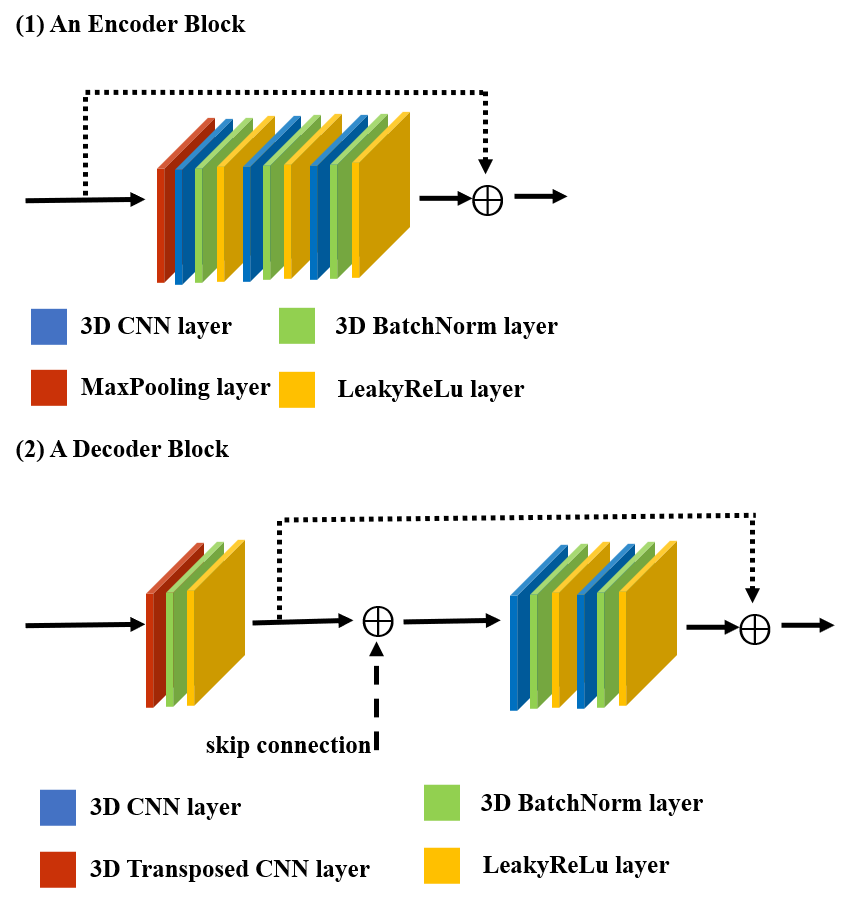}
\end{center}
   \caption{Architecture of the encoder block and the decoder block. The dotted lines show residual connections. The dash lines represent the skip connections from the encoder to the decoder. }
\label{fig:short}
\end{figure}

 \begin{figure*}
\begin{center}
\includegraphics[width=0.9\linewidth]{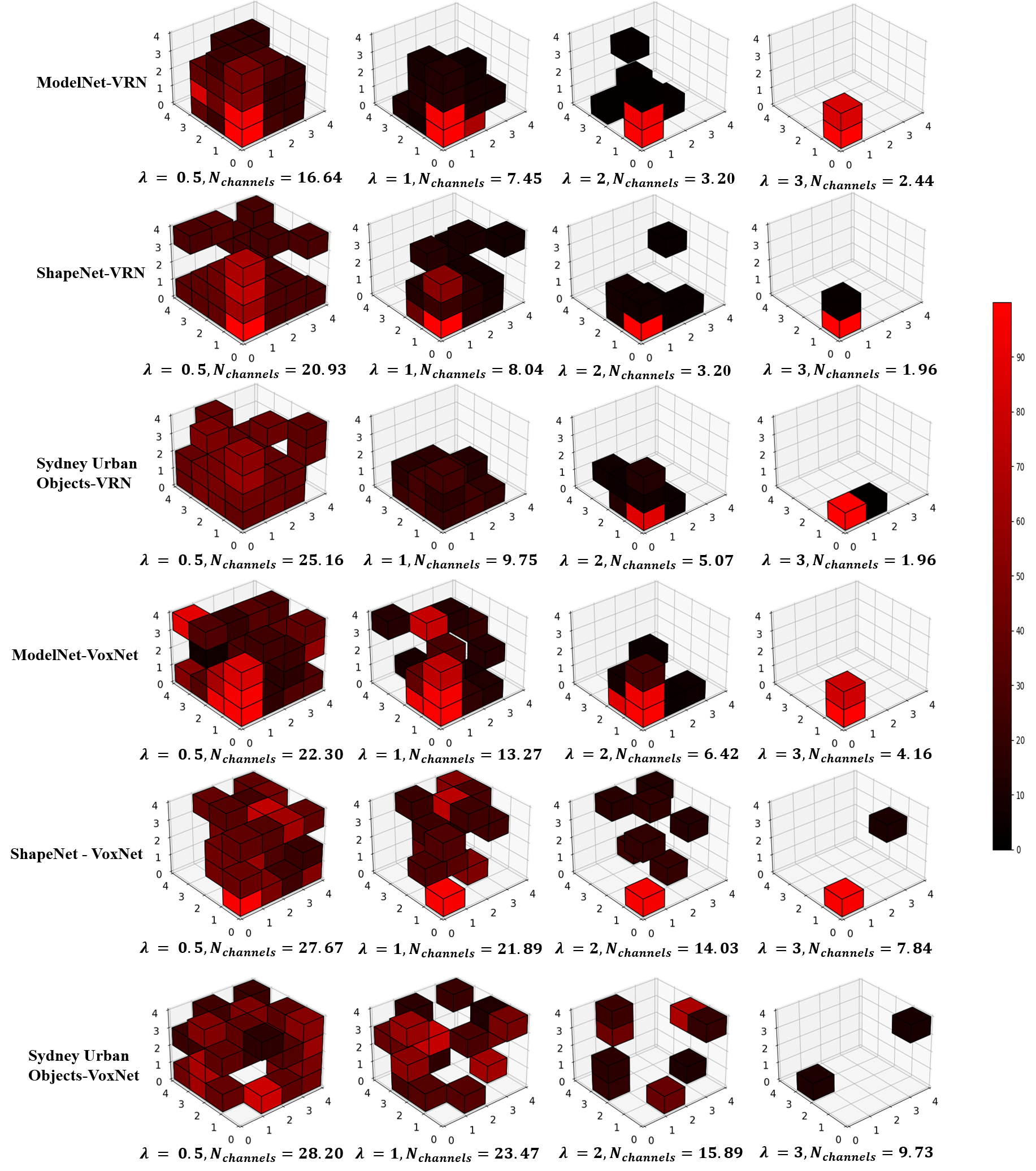}
\end{center}
   \caption{Heat maps of top selected frequency channels in shape classification datasets by using VRN and VoxNet. The heat map value means the likelihood that a channel is to be activated. $N_{channels}$ means the average number of selected channels by a sample during dynamic training. We show top $32$ frequency channels when $\lambda = 0.5$, top $16$ frequency channels when $\lambda = 1$, top $8$ frequency channels when $\lambda = 2$ and top$2$ frequency channels when $\lambda = 3$. }
\label{fig:short}
\end{figure*}

As the size of input data in the frequency domain is smaller than the size of its corresponding data in the spatial domain, we avoid downsampling on inputs in the frequency domain and keep the network architecture unchanged. We expand the number of filters in the hidden layers to keep a comparable amount of features with the spatial-domain inputs for the 3D CNN model with few layers. For example, the VoxNet \cite{voxnet}, which is one of the baselines we used for shape classification, has two 3D convolutional layers and one pooling layer. We set the stride of the two convolutional layers to 1 and the number of filters to 128. Fig. 3 shows the architecture comparison of our spatial-domain VoxNet and frequency-domain VoxNet, which have the same number of layers but the different number of filters in 3D CNN layer. For deeper networks, i.e. VRN \cite{vrn}, we remove the downsampling blocks in the early stage until the output of a hidden layer with spatial-domain input data has the close size of its corresponding data in the frequency domain. We keep the following blocks all the same with the network dealing with spatial-domain inputs. The VRN network \cite{vrn} has four blocks, and one block contains a VRN block and a downsample block. The size of the second block's output is $8 \times 8 \times 8$, which is the same size as the inputs in the frequency domain. Hence, we remove the downsampling part and double the number of filters in the first two blocks to keep a similar feature amount at each hidden layer. Fig. 4 presents the architecture of our frequency-domain VRN. In the architecture of the original VRN, Block 1 and Block 2 also have downsampling (DS) blocks. 

In terms of point cloud segmentation, we implement an encoder-decoder style fully convolutional network as our baseline method. Fig. 5 shows the architectures of our spatial-domain and frequency-domain point cloud segmentation methods. The encoder part has four fully CNN encoder blocks and yields a $16\times$ downsampling. The decoder part has four fully CNN decoder blocks. Inspired by the encoder-decoder style networks with skip connections in 2D image segmentation tasks, we add the four skip connections from the encoder part to the decoder part to provide texture information lost during downsampling. We use the dash lines to show the four skip connections in fig. 5. Fig. 6 presents the architecture of an encoder block and a decoder block. An encoder block first downsamples the input by a 3D max pooling layer and extracts features by three 3D CNN layer blocks with $kernal \: size = 3$ and $stride = 1$. One 3D CNN layer block contains one 3D CNN layer, one Leaky ReLu layer, and one batch normalization layer. A decoder block first upsamples the input feature maps by a 3D transposed CNN layer with $kernal \: size = 4$ and $stride = 2$. Then, the upsampled feature maps are merged with feature maps from the encoder part by element-wise addition. Two 3D CNN blocks follow the 3D transposed CNN layer to extract features. The dotted line in the fig. 6 stands for a residual connection inspired by the ResNet \cite{resnet}.  For frequency-domain learning, we remove the max pooling layers in the first two encoder blocks to avoid downsampling and upsample the outputs of the first two 3D encoder blocks to provide the skip connections for the decoder. The trilinear interpolation layer transfers the voxel-level predictions to point-level predictions, which reduce the false predictions caused by the low-resolution 3D volumetric grid. 

 \begin{figure*}
\begin{center}
\includegraphics[width=0.9\linewidth]{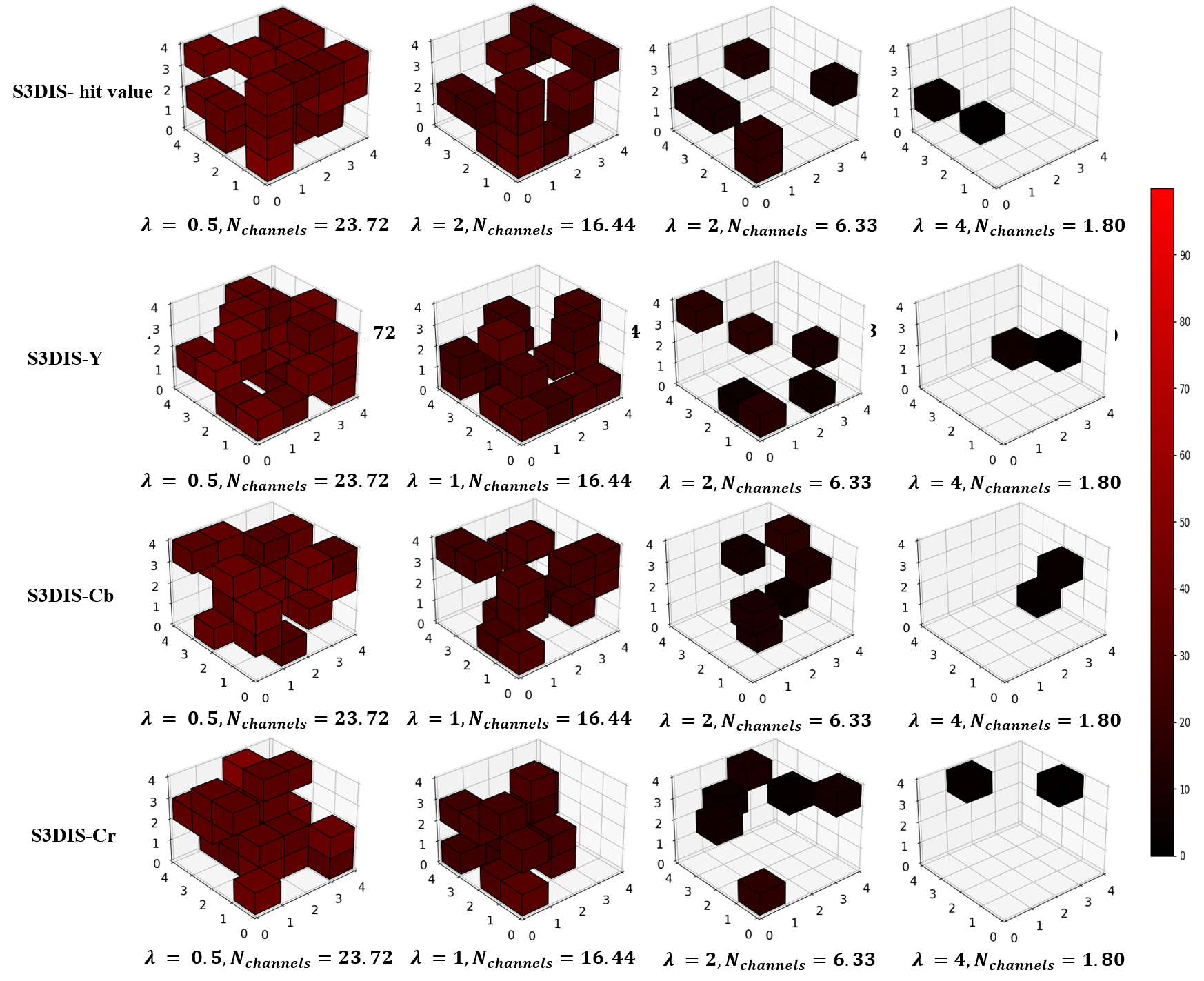}
\end{center}
   \caption{Heat maps of top selected frequency channels in S3DIS dataset by using the encoder-decoder style baseline method. $N_{channels}$ means the average number of selected channels from each attribute by a sample during dynamic training.}
\label{fig:short}
\end{figure*}

\subsection{Static Frequency Channel Selection for 3D Volumetric Data}
 
\subsubsection{Spectral Bias of 3D CNN models}
As shown in Fig. 7, we plot heat maps to demonstrate the spectral bias of 3D CNN models in shape classification. The heat map shows that the likelihood that a frequency channel being activated by the gate module during inference. Elements in a heat map stand for frequency channels, and the axes indicate the coordinate of the frequency channels, e.g., the coordinate of the DC frequency channel is (0, 0, 0). The color of elements indicates the activation probabilities. The $\lambda$ is the hyperparameter term in Equation (2). $N_{channels}$ is the average number of selected channels for each sample. When the $\lambda$ increases, $N_{channels}$ will decrease as the method tends to activate fewer frequency channels to reduce the loss in the joint training. We only show top frequency components in each heat map for visualization.

For shape classification datasets, the ModelNet \cite{modelnet} and ShapeNet \cite{shapenet} contain CAD models of man-made objects, and the Sydney Urban Objects dataset contains point clouds of outdoor objects. Their heat maps indicate that in most cases, the low-frequency channels, especially the DC component channel, have higher probabilities to be selected when $\lambda = 3$.  As the number of activated channels decrease, only the DC component has probabilities over $99\%$ to be activated. In the rare case of the spectral bias of VoxNet with the Sydney Urban Objects dataset, no frequency channels have noticeable higher probabilities to be selected, when the average number of selected channels decreases, which indicates that spectral bias are variant for distinct CNN models and dataset. However, in general, most of the activated frequency channels are low-frequency channels or have a low-frequency attribute along at least one dimension in space. Compared to the Voxnet with simple network architecture, the deeper the 3D CNN model, i.e. VRN, the fewer frequency channels are selected under the same $\lambda$ in the same dataset. By observing heat maps in the same dataset and same method, we noticed that the frequency channels with top activation probabilities are stable under different $\lambda$. 

The spectral bias of the point cloud segmentation task on S3DIS is shown in Fig. 8 and shows different insights from the spectral bias in shape classification. The top 3D frequency channels that are most informative to the point cloud semantic segmentation are more distributed across the spectrum. The activation likelihood of top frequency channels in the heat maps of $\lambda = 0.5$ decreases as the number of activated channels decreases, and no dominated frequency channels like the DC frequency channel in the shape classification have a higher likelihood (e.g., over 80\%) to be activated and most frequency channels have close and low probabilities to be activated when $\lambda = 4$.  However, most of the top selected channels have least a low-frequency attribute along a dimension in space under each condition, and the results of our following experiments show that reserving the frequency channels with top probability is enough to obtain comparable accuracy and IoU in the point cloud segmentation task as compared with spatial-domain approaches with full-size data.

\subsubsection{Static Frequency Channel Selection}
Based on the above observations, we statically select the top frequency channels based on the rank of the probabilities to be activated and train the models from scratch with the selected frequency channels to explore whether a deterministic frequency channel selection method is generic for all samples. 

Although the top frequency channels are stable under different $\lambda$, some frequency channels, e.g. the DC frequency channel in the spectral bias of VRN with Sydney urban objects dataset, show the dominance when $\lambda$ is large enough. In our method, we select top $n$ frequency channels from the spectral bias when $\lambda = 2$ for static channel selection training, and $n$ should be smaller than the average number of selected channels in spectral bias. And then, we train the frequency-domain networks without the gate module on $n$ selected channels. For inference, we use a deterministic binary channel selection map for all test samples, as shown in Fig. 1, to select frequency channels of each test data, where '1' means channels to be selected, and '0' means muted channels. Our experiments results verify our hypothesis that models trained by statically selected frequency channels are able to achieve comparable results with full-size data in the spatial domain. Fig. 9 shows the trade-off between accuracy and normalized input size in inference on the pre-trained models. The models trained by selected frequency channels achieves comparable results with smaller normalized input size, which demonstrate most of the frequency channels especially high-frequency channels are redundant in 3D shape classification and 3D point cloud segmentation.

\section{Experiments}

\subsection{3D Shape Classification}

\begin{figure}
\begin{center}
   \includegraphics[width=\linewidth]{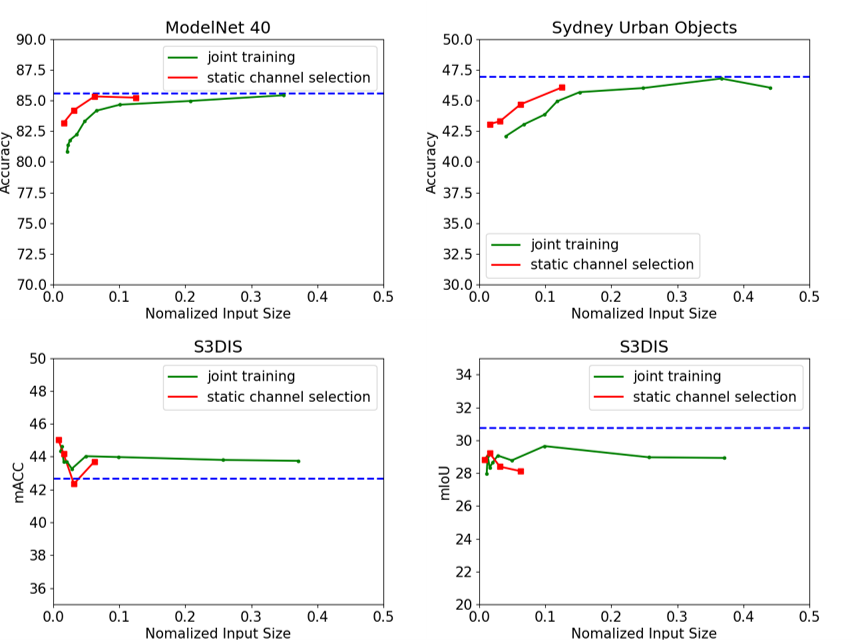}
\end{center}
   \caption{Tradeoff between accuracy and normalized input data size. The blue dotted line in the above four pictures are the accuracy or meanIoU of the models trained by full-size spatial-domain volumetric data. 'Joint training' is the results of the models with the gate module trained by full-size frequency-domain data. 'Static channel selection' is the results of the models trained by selected frequency channels.}
\label{fig:long}
\label{fig:onecol}
\end{figure}

\subsubsection{Experiment Setup}
We explore 3D frequency-domain learning in 3D shape classification using ModelNet dataset \cite{modelnet}, ShapeNet dataset \cite{shapenet} and Sydney Urban Object dataset. The ModelNet \cite{modelnet} contains CAD models of most common objects of size $32 \times 32 \times 32 $. Each sample is augmented by rotating in 12 directions, so our observations of spectral bias are rotation invariant. We utilize a subset with 10 categories and a subset with 40 categories separately. The ShapeNet \cite{shapenet} includes about 51,300 unique 3D models with 55 categories. It provides the volumetric data of size $128 \times 128 \times 128$, and we downsample it to $32 \times 32 \times 32$. Both surface-volumetric data and filled-in volumetric data are utilized for our experiments. Sydney Urban Object dataset contains 631 urban objects in 26 categories, which are cropped from LiDAR scans. We followed the provided dataset split method, which covers 15 classes and some classes with less samples are dropped. We represent each object by a volumetric grid of $32 \times 32 \times 32$. 

We train VoxNet for 32 epochs on the above mentioned three datasets and decay the learning rate by 0.8 every 4 epochs. We use the stochastic gradient descent (SGD) optimizer with the an initial learning rate of $0.001$, a momentum of $0.9$, and a weight decay of $0.001$. We use the simplest VRN \cite{vrn} network instead of an ensemble of different VRN models. For the VRN, we train it for 50 epochs on all the same datasets except for the ModelNet-10, as \cite{modelnet} claims that their proposed network does not obtain considerable accuracy on ModelNet-10 as lack of data.  The initial learning rate of SGD is $0.002$, and the learn rate decay by $0.1$ when the loss of a validation set stop reducing. The other settings keep the same with VoxNet \cite{voxnet}. The loss function of the above two networks is cross-entropy loss with the regularization term mentioned in Section 3.2. 

\subsubsection{Experiment Results}
We train the original VoxNet and VRN with volumetric data of size $1 \times 32 \times 32 \times 32$. Then, We train the frequency-domain VoxNet \cite{voxnet} and VRN \cite{vrn} with the gate module using the 64 channel volumetric data in the frequency-domain to explore the variation of classification accuracy with dynamically selected frequency channels by comparing with the results of original methods. We plot heat maps to demonstrate the probability statistics of activated frequency channels (see Fig. 7). The heat maps show the activation probability of top $n$ frequency channels. When $\lambda = 3$ in 3D shape classification datasets, at most two frequency channels have activation probability over 80\%. Then, we utilize the static frequency channel selection method mention in section 3.4 to train frequency-domain VoxNet \cite{voxnet} and VRN \cite{vrn} with selected frequency channels. In most instances, when only one frequency channel is selected, the selected frequency channel is the DC frequency component channel. 

\begin{table}
\begin{center}
\scalebox{0.75}{
\begin{tabular}{|l|c|c|}
\hline
VoxNet & Normalized Input Size & Acc \\
\hline\hline
Voxel(binary) & $ 1.0 $ & 90.07\\
DCT $N_{channels} = 64$ & 1.0 & 89.29\\
DCT $N_{channels} = 4$ & 0.063 & 89.49 \\
DCT $N_{channels} = 2$ & 0.031 & 90.23\\
DCT $N_{channels} = 1$ & 0.016 & \textbf{91.00}\\
\hline
\end{tabular}}
\end{center}
\caption{VoxNet classification results on ModelNet-10. ``Voxel(binary)" means the full-size data in the spatial domain by using the binary hit grid occupancy model to quantify voxels, and ``DCT" means data in the frequency domain. ``$N_{channels}$" is the number of activated channels. }
\end{table}\textbf{
}

\begin{table}
\begin{center}
\scalebox{0.75}{
\begin{tabular}{|l||l|l|l|}
\hline
Network                 & Data & Normalized Input Size & Acc \\ \hline\hline
VoxNet & Voxel(binary) & 1.0 & 85.59\\ \cline{2-4} 
                        &DCT $N_{channels}=64$ & 1.0 & 85.55 \\ \cline{2-4} 
                        &DCT  $N_{channels}=4$ & 0.063 & \textbf{85.33}\\ \cline{2-4} 
                        &DCT  $N_{channels}=2$ & 0.031 & 84.21\\ \cline{2-4} 
                        &DCT  $N_{channels}=1$ & 0.016 & 83.17 \\ \hline
VRN    &  Voxel(binary) & 1.0 & 88.11 \\ \cline{2-4} 
    & DCT $N_{channels}=4$ & 0.063 & \textbf{85.13} \\ \cline{2-4} 
    & DCT $N_{channels}=2$ & 0.031 & 84.56 \\ \cline{2-4} 
    & DCT $N_{channels}=1$ & 0.016 & 83.42 \\ \hline
\end{tabular}}
\end{center}
\caption{classification results on ModelNet-40 }
\end{table}

The accuracy of ModelNet-10 outperforms the baseline method with full-size data in the spatial domain at most $0.93\%$, and the accuracy of ShapeNet surface volumertric data outperforms the baseline method with full-size data in the spatial domain at most $0.58\%$.  The results are consistent with our observed spectral bias that the DC component channel contains most of important information. Note that in some cases (i.e., ModelNet-10 and ShapeNet), the accuracy drops when the number of activated channels increases because of the noisy information in high-frequency channels. Additionally, note that the accuracy of the experiment with the 64 channel full-size frequency domain data is below than the full-size data in the spatial domain. We conjecture the accuracy gap between full-size frequency-domain data and spatial-domain data implies the drawback of CNN models to deal with frequency-domain data. For experiments performed using the VRN \cite{vrn} network (see Table 2-5), the overall accuracy is higher than the VoxNet \cite{voxnet} with data in the frequency domain. The models trained by one selected channel achieve comparable accuracy with full-size data in the spatial domain but the normalized input data size is smaller, e.g., $-2.04\%$ on Sydney Urban Objects dataset and $-2.03\%$ on ShapeNet surface volumetric data but the normalized input size is only 0.0156. When 2 frequency channels are selected, compared to full-size data in the spatial domain, the accuracy of dimensionality reduced data drops about $1.96\%$ on ShapeNet surface volumetric data and $0.32\%$ on Sydney Urban Objects with a 0.031 normalized input size .

\begin{table}
\begin{center}
\scalebox{0.75}{
\begin{tabular}{|l||l|l|l|}
\hline
Network                 & Data & Normalized Input Size & Acc \\ \hline\hline
VoxNet & Voxel(binary) & 1.0 & 46.97\\ \cline{2-4} 
                        &DCT $N_{channels}=4$ & 0.063 & \textbf{44.68} \\ \cline{2-4} 
                        &DCT $N_{channels}=2$ & 0.031 & 43.30\\ \cline{2-4} 
                        &DCT $N_{channels}=1$ & 0.016 & 43.08\\ \hline
VRN    & Voxel(binary) & 1.0 & 48.86\\ \cline{2-4} 
    & DCT $N_{channels}=4$ & 0.063 & 47.48\\ \cline{2-4} 
    & DCT $N_{channels}=2$ & 0.031 & \textbf{48.54} \\ \cline{2-4} 
    & DCT $N_{channels}=1$ & 0.016  & 46.82 \\ \hline
\end{tabular}}
\end{center}
\caption{classification results on Sydney Urban Objects }
\end{table}

\begin{table}
\begin{center}
\scalebox{0.75}{
\begin{tabular}{|l||l|l|l|}
\hline
Network                 & Data & Normalized Input Size & Acc \\ \hline\hline
VoxNet & Voxel(binary) & 1.0 & 81.58\\ \cline{2-4} 
                        &DCT $N_{channels}=4$ & 0.063 & \textbf{82.01} \\ \cline{2-4} 
                        &DCT $N_{channels}=2$ & 0.031& 81.93\\ \cline{2-4} 
                        &DCT $N_{channels}=1$ & 0.016 & 82.16\\ \hline
VRN    & Voxel(binary) & 1.0& 85.59\\ \cline{2-4} 
    & DCT $N_{channels}=4$ & 0.063 & \textbf{83.85}\\ \cline{2-4} 
    & DCT $N_{channels}=2$ & 0.031& 83.63 \\ \cline{2-4} 
    & DCT $N_{channels}=1$ & 0.016 & 83.56 \\ \hline
\end{tabular}}
\end{center}
\caption{classification results on ShapeNet surface volumetric data }
\end{table}

\begin{table}
\begin{center}
\scalebox{0.75}{
\begin{tabular}{|l||l|l|l|}
\hline
Network                 & Data & Normalized Input Size & Acc \\ \hline\hline
VoxNet &Voxel(binary) & 1.0 & 82.10\\ \cline{2-4} 
                        &DCT $N_{channels}=4$ & 0.063 & \textbf{82.26} \\ \cline{2-4} 
                        &DCT $N_{channels}=2$ & 0.031 & 81.42\\ \cline{2-4} 
                        &DCT $N_{channels}=1$ & 0.016& 81.48\\ \hline
VRN    & Voxel(binary) & 1.0 & 86.62\\ \cline{2-4} 
    & DCT $N_{channels}=4$ & 0.063 & \textbf{83.36}\\ \cline{2-4} 
    & DCT $N_{channels}=2$ & 0.031 & 82.64 \\ \cline{2-4} 
    & DCT $N_{channels}=1$ & 0.016 & 82.26 \\ \hline
\end{tabular}}
\end{center}
\caption{classification results on ShapeNet filled-in volumetric data }
\end{table}

\subsection{Point Cloud Segmentation}

\subsubsection{Experiment Setup}

We explore 3D frequency-domain learning in 3D point cloud segmentation using using Stanford Large-Scale 3D Indoor Spaces dataset (S3DIS). \cite{S3DIS} contains large scale 3D points cloud for six reconstructed areas. The dataset covers 14 common indoor categories, and 13 among them used for evaluation.  We used the same dataset settings with \cite{segcloud}, which tests their method on the fifth area, and trains their method on the rest.

One area of the S3DIS \cite{S3DIS} contains multiple scenarios. To keep a higher resolution of volumetric data, we divide each scenarios into several $5m \times 5m \times 5m$ sub-blocks. Each sub-block is voxelized in a grid of size $80 \times 80 \times 80 $. A voxel has a resolution of $0.0625 m$. For the experiments with higher spatial-domain resolution volumetric data, we voxelized each sub-block in a grid of size $160 \times 160 \times 160$, and a voxel has a resolution of $0.03125 m$.  One voxel has four attributes - binary occupation state, Y, Cb, and Cr. The color attribute of a voxel is the average value of each color information channel of all points in a voxel. We train the encoder-decoder-style fully convolutional model with 200 epochs and decay the learning rate by 0.1 every 50 epochs. We use an Adam optimizer with an initial learning rate of 0.001 and a weight decay of 0.001. The loss function is a point-wise cross-entropy loss with class weights. 

\subsubsection{Experiment Results}

\begin{table*}[t]
\begin{center}
\resizebox{\linewidth}{!}{
\begin{tabular}{|l|l|l|l|l|l|l|l|l|l|l|l|l|l||l|l|l|}
\hline
 & door & floor & wall & ceiling & clutter & chair & board & bookcase & beam & table &window & sofa & column  & Acc & mAcc & mIoU  \\
\hline\hline
Voxel + RGB & \textbf{15.87} & \textbf{94.47} & \textbf{53.43} & 73.26 & \textbf{17.38} & \textbf{25.23} & 7.39 & \textbf{37.49} & 0.26 & \textbf{31.22} & 31.68 & 2.22 & 10.09 & \textbf{87.33} & 42.69 & \textbf{30.77}\\
\hline
 DCT $N_{channels} = 4$ & 12.75 & 94.09 & 34.65 & 75.69 & 14.41 & 19.23 &
 11.50 & 36.56 & 0.00 & 26.71 & 38.79 & \textbf{4.72} &
 \textbf{10.71} & 84.73 & 44.17 & 29.22 \\
 \hline
DCT $N_{channels} = 2$ & 11.90 & 93.31 & 30.69 & \textbf{76.97} & 10.27 & 23.93 & \textbf{12.23} & 33.92 & 0.00  & 30.87 & \textbf{39.60} & 1.45 & 9.63 & 84.15 & \textbf{45.02} & 28.83\\ 
\hline
\end{tabular}
}
\end{center}
\caption{Point segmentation results of the model trained by full-size data in the spatial domain and selected frequency channels}
\end{table*}

\begin{table}
\begin{center}
\scalebox{0.75}{
\begin{tabular}{|l|c|c|c|c|}
\hline
 & Normalized Input Size & ACC & mACC & mIoU \\
\hline\hline
Voxel+RGB & $ 1.0 $ & 87.33 & 42.69 & 30.77\\
LR DCT $N_{channels} = 2$ & 0.008 & 84.15 & 45.02 & 28.83\\
LR DCT $N_{channels} = 4$ & 0.016 & 84.73 & 44.17 & 29.22 \\
HR DCT $N_{channels} = 2$ & 0.063 & 85.01 & 43.67 & 29.47\\
HR DCT $N_{channels} = 4$ & 0.125 & 86.28 & 45.73 & 31.40\\
\hline
\end{tabular}
}
\end{center}
\caption{Point cloud segmentation results of models trained by selected high-resolution frequency channels. "HR" is the high-resolution data. "LR" is the low-resolution data.}
\end{table}

We show quantitative results of trained models by full-size data in the spatial domain and the selected frequency channels separately in Table 6. We evaluate the models by the overall classification accuracy of all points, mean of classification accuracy of all classes (mAcc) and mean of intersection over union of all classes (mIOU). For the model trained by two selected frequency channels, we select one frequency channel from the voxel binary occupancy state attribute and one frequency channel from the Y attribute, and no channel is selected from Cb and Cr attributes. For other models trained by selected frequency channels, the channels are selected averagely from all attributes. In most cases, the models trained by selected frequency channels outperform the baseline method using full-size spatial domain data at most $2.33\%$ in terms of mACC and loses at most $2.65\%$ on mIoU. The experiment results of models trained by selected frequency channels from high-resolution transformed data (see Table 7) present that the frequency-domain learning model outperforms the full-size spatial-domain learning model at $0.63\%$ on mIoU and $3.04\%$ on mACC with an 87.5\% smaller normalized input size than full-size low-resolution spatial-domain input. Figure 10 shows the inference visualization results of point cloud segmentation with models trained by low-resolution frequency channels.

As the frequency-domain input size is smaller than the corresponding spatial domain input size, the memory footprint and FLOPs are reduced in the first two encoder blocks of our point cloud segmentation baseline. For the high-resolution spatial-domain input of size $160 \times 160 \times 160$, the FLOPs needed in inference is about 1127 GFLOPs. The model that takes frequency-domain input needs about 1025 GFLOPs, and the frequency-domain learning reduces about 9\% FLOPs. On the other hand, the GPU memory footprint requirement for a spatial-domain input of size $160 \times 160 \times 160$ is about 8047 MB in inference, and the corresponding frequency-domain input occupies about 1632 MB which means about 20\% GPU memory footprint requirement is reduced. 

\begin{figure}[t]
\begin{center}
   \includegraphics[width=\linewidth]{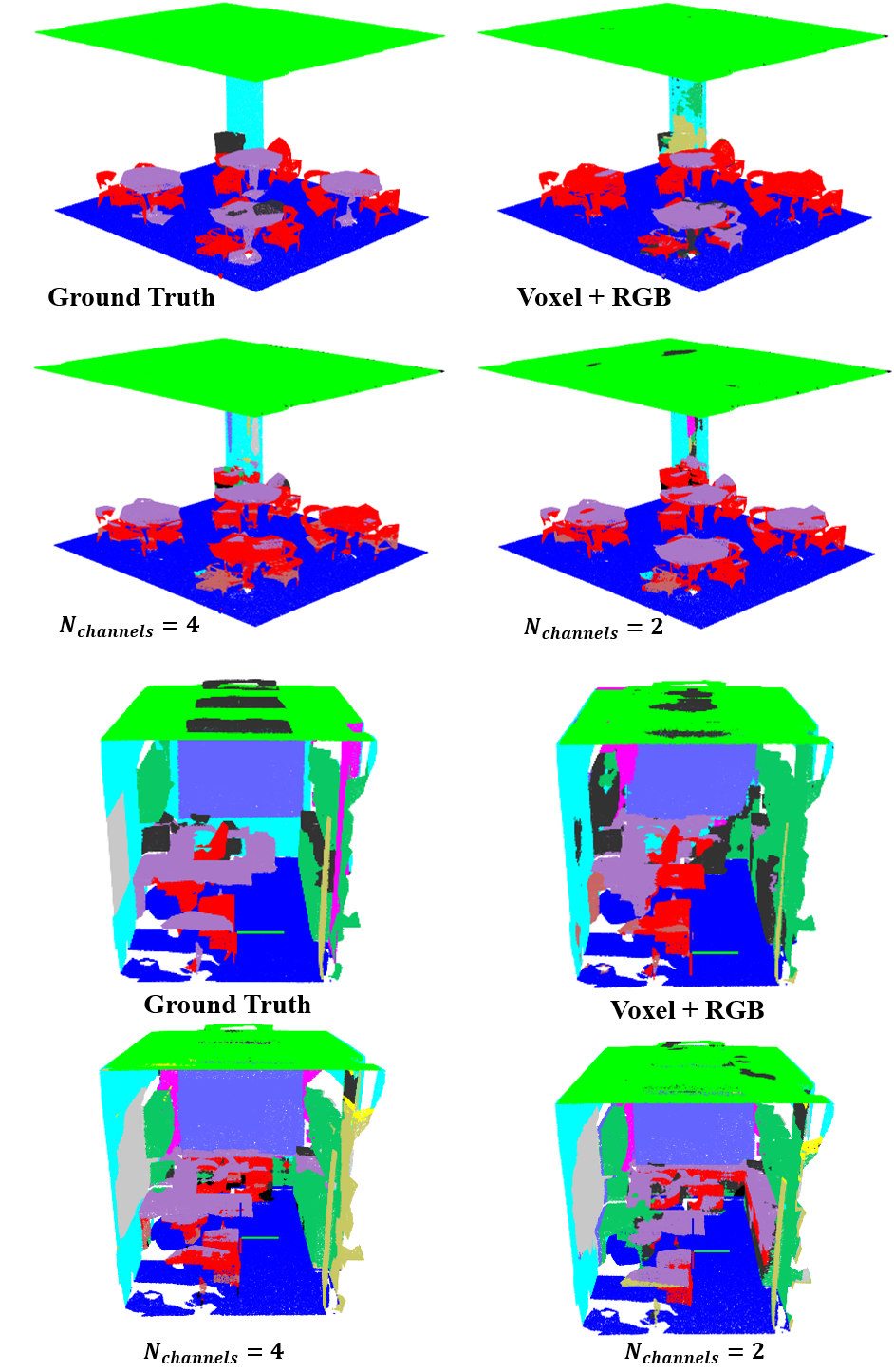}
\end{center}
   \caption{Examples of point cloud segmentation results on low-resolution data, i.e. $80 \times 80 \times 80$, of S3DIS.}
\label{fig:long}
\label{fig:onecol}
\end{figure}

\section{Conclusion}
In this paper, we study frequency-domain learning for volumetric-based 3D data perception. Our 3D frequency-domain learning utilizes existing 3D CNN models with little modifications to reveal the spectral bias of 3D CNN models. Experiment results of the static frequency channel selection method demonstrate that only frequency channels with high activation probability, e.g., over 90\%, contain key information for volumetric-based 3D data perception. Experiment results show that frequency-domain learning is able to reduce the input data size by 98\% in 3D shape classification while limiting the average accuracy drop within 2\%, and by 98\% in 3D point cloud semantic segmentation with a 1.48\% mean-class accuracy improvement while limiting the mean-class IoU loss within 1.55\%. Our experiment results also indicate that 3D frequency-domain learning can effectively reduce the input data size when learning from high-resolution volumetric data, which potentially alleviates the large requirements for computation resources and GPU memory footprints in volumetric-based 3D vision tasks and avoid complicated modification to the existing 3D CNN models. 


\begin{thebibliography}{}
%
%
\bibitem{2dfdl}
Kai Xu, Minghai Qin,  Fei Sun, Yuhao Wang, Yen kuangChen, and Fengbo Ren. Learning in the frequency domain. \textit{IEEE Conf. Comput. Vis. Pattern Recog.}, 2020.
\bibitem{3dsurvey}
Yulan Guo, Hanyun Wang, Qingyong Hu, Hao Liu, Li Liu,and  Mohammed  Bennamoun. Deep learning for 3d pointclouds: A survey. \textit{IEEE Transactions on Pattern Analysisand Machine Intelligence}, 2020.
\bibitem{voxnet}
Daniel Maturana and Sebastian Scherer. Voxnet: A 3d con-volutional neural network for real-time object recognition. \textit{IEEE/RSJ International Conference on Intelligent Robotsand Systems(IROS)}, 2015.
\bibitem{vrn}
Andrew Brock, Theodore Lim, J.M. Ritchie, and Nick We-ston.    Generative and discriminative voxel modeling with convolutional neural networks.\textit{Adv. Neural Inform. Process. Syst.}, 2016.
\bibitem{jpeglearning}
Lionel Gueguen, Alex Sergeev, Ben Kadlec, Rosanne Liu, and Jason Yosinski. Faster neural networks straight from jpeg. \textit{Adv. Neural Inform.  rocess. Syst.}, 2018.
\bibitem{resnet}
Charles R. Qi., Hao Su, Kaichun Mo, and Leonidas J. Guibas. Deep residual learning for image recognition. \textit{IEEEConf. Comput. Vis. Pattern Recog.}, 2016.
\bibitem{jpegdetection}
Benjamin Deguerre, Clement Chatelain, and Gilles Gasso. Fast object detection in compressed jpeg images. \textit{IEEE Intelligent Transportation Systems Conference}, 2019.
\bibitem{subvolume}
Charles R. Qi, Hao Su, Matthias Nießner, Angela Dai, Mengyuan Yan, and Leonidas J. Guibas. Volumetric and multi-view cnns for object classification on 3d data. \textit{IEEEConf. Comput. Vis. Pattern Recog.}, 2016.
\bibitem{inception}
Christian Szegedy, Wei Liu, Yangqing jia, Pierre Sermanet, Scott  Reed, Dragomir Anguelov, Dumitru Erhan, Vincent Vanhoucke, and Andrew Rabinovich. Going deeper with convolutions. \textit{IEEE Conf. Comput. Vis. Pattern Recog.}, 2015.
\bibitem{pointseg1}
ing Huang and Suya You. Point cloud labeling using 3d convolutional neural network.\textit{Int. Conf. Pattern Recog.}, 2016.
\bibitem{segcloud}
Lyne P. Tchapmi, Christopher B. Choy, Iro Armeni, JunY-oung Gwak, and Silvio Savarese. Segcloud: Semantic segmentation of 3d point clouds. \textit{Intenational Conference on 3D Vision}, 2017.
\bibitem{rangenet}
Andres Milioto, Ignacio Vizzo, Jens Behley, and CyrillStachniss. Rangenet++: Fast and accurate lidar semantic segmentation. \textit{IEEE/RSJ International Conference on Intelligent Robots and Systems (IROS)}, 2019.
\bibitem{modelnet}
Zhirong Wu, Shuran Song, Aditya Khosla, Fisher Yu, Lin-guang Zhang, Xiaoou Tang, and Jianxiong Xiao. 3dshapenets: A deep representation for volumetric shapes. \textit{IEEE Conf. Comput. Vis. Pattern Recog.}, 2015.
\bibitem{shapenet}
Angel X. Chang, Thomas Funkhouser, Leonidas Guibas, PatHanrahan, Qixing Huang, Zimo Li, Silvio Savarese, Mano-lis Savva, Shuran Song, Hao Su, Jianxiong Xiao, Li Yi, and Fisher Yu. Shapenet: An information-rich 3d model repository. Technical Report arXiv:1512.03012 [cs.GR], StanfordUniversity — Princeton University — Toyota Technological Institute at Chicago, 2015.
\bibitem{S3DIS}
Iro Armeni, Ozan Sener, Amir R. Zamir, Helen Jiang, Ioannis Brilakis, Martin Fischer, and Silvio Savarese. 3d semantic parsing of large-scale indoor spaces. \textit{IEEE Conf. Comput.Vis. Pattern Recog.}, page 1534–1543, 2016.
\end{thebibliography}

\end{document}